%% file: main.tex
\pdfoutput = 1
\documentclass[twocolumn]{article}
\input{macros}

\usepackage{style}

\makeatletter
\@namedef{ver@everyshi.sty}{}
\newcommand{\removelatexerror}{\let\@latex@error\@gobble}

\title{Constrained Sampling for Class-Agnostic Weakly Supervised Object Localization}

\renewcommand\footnotemark{}

\author{Shakeeb Murtaza$^{1}$,
Soufiane~Belharbi$^{1}$,
  ~Marco~Pedersoli$^{1}$, 
  ~Aydin~Sarraf$^{2}$,  and
  ~Eric~Granger$^{1}$\\
 	$^1$ LIVIA, Dept. of Systems Engineering, ÉTS, Montreal, Canada \\
	$^2$ Ericsson, Global AI Accelerator, Montreal, Canada\\
{\tt\footnotesize \textcolor{black}{shakeeb.murtaza.1@ens.etsmtl.ca} }
}

\newcommand{\ignore}[1]{}

\begin{document}
\maketitle\thispagestyle{fancy}

\maketitle
\lhead{\color{gray} \small \today}
\rhead{\color{gray} \small Murtaza et al. \;  [
Montreal AI Symposium 2022]}


\textbf{Code:} \href{https://github.com/shakeebmurtaza/dips}{https://github.com/shakeebmurtaza/dips}.
% ===================================================================================================
%
%                                      INTRODUCTION
%
% ===================================================================================================

%%%%%%%%% BODY TEXT
\section{Introduction}
\label{sec:intro}
Weakly supervised learning (WSL) is a paradigm that allows training machine learning models from datasets with incomplete, inexact, and ambiguous labels~\citep{zhou2018brief}. In computer vision, WSL has received much recent attention for the object localization task, where training typically requires costly pixel-wise annotations. In particularly, weakly supervised object localization (WSOL) involved training ML models to localize objects of interest using data with only image-class labels. Class Activation Mapping (CAM) methods are a popular method for WSOL, allowing to highlight image regions linked to a particular class using features from the penultimate layer~\citep{zhou2016learning}. However, CAMs provides coarse information on object location for a given input image, often resulting in poor localization performance because they tend to focus on the most discriminative regions shared among objects of the same class~\citep{negevsbelharbi2022,belharbi2022f}.  Moreover, CAM methods produce continuous activation maps that may be affected by complex background regions that match the concerned (foreground) object. Therefore, optimal thresholds must be selected to produce an accurate bounding box.

\begin{figure*}[!htp]
\begin{center}
\includegraphics[width=\linewidth,trim=3 3 3 3, clip]{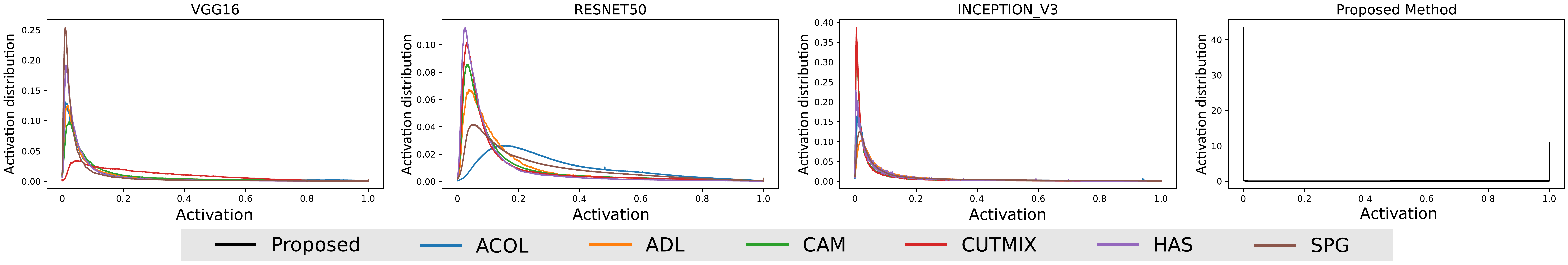}
\end{center}
\vspace{-2em}
\caption{Distribution of activation scores for CUB test set at different thresholds.\vspace{-2em}}\label{fig:dist_shift}
\end{figure*}

\section{Proposed Approach}
\label{sec:proposal}

To address the aforementioned issues with encoder-decoder networks, we propose a method for constrained sampling of pixels from foreground and background regions to construct an efficient pseudo-label by employing attention maps generated by self-supervised transformers. These pseudo-labels allow to directly minimize standard cross-entropy loss for both foreground and background regions, thus producing a high-resolution map with the same intensities all over the object and does not require an optimal threshold for each image. Moreover, the generated map is class-agnostic and only contains two channels, one for background and the other for foreground regions. Furthermore, the localization map predictor is trained using pixel-level pseudo labels extracted through our proposed method. It samples pseudo-pixels from attention maps of the last transformer layer trained in a self-supervised way, corresponding to \textit{[cls]} tokens. From these maps, only the first four maps are selected along with an average of all attention maps. Then, the Otsu \citep{otsu1979threshold} thresholding method is applied to the selected maps for converting them into a binary map and find all possible $\textbf{n}$ bounding boxes for each of the five maps. Since each of these attention maps may highlight a different object, we can infer one bounding box for every object. Then, given an input image, $\textbf{n}$ images are produced, corresponding to each bounding box by applying the Gaussian-blur filter to hide the background regions outside of the respective bounding box. Each image is then passed to a classifier ${\mathrm{C}_\theta}$ to select the best proposal, along with its underlying attention map $T_i$. This corresponds to the best classifier's confidence and underlying target class $\mathrm{y}_i$. Finally, the pixels inside and outside of the bounding box are considered as foreground regions ${\mathbb{Y}_i^+}$ and background regions ${\mathbb{Y}_i^-}$. Here, ${\mathbb{Y}_i^+}$ represents the top $n\%$ pixels and ${\mathbb{Y}_i^-}$ represents top $-n\%$ pixels derived from activations that is sorted in reverse order. Additionally, class and conditional random field (CRF) loss may also be employed to validate the robustness of maps and to align them with the object's boundaries. These auxiliary losses are employed because pseudo labels may contain imprecise information regarding object regions. Moreover, these losses ensure the alignment of the map with the actual object as we are selecting a few pixels for pseudo labels.

%%%%%%%%%%%
\subsection{Experimental Results}
\label{subsec:exps}

\begin{figure*}[!htp]
\centering
\includegraphics[width=1\linewidth,trim=0 2320 0 0,clip]{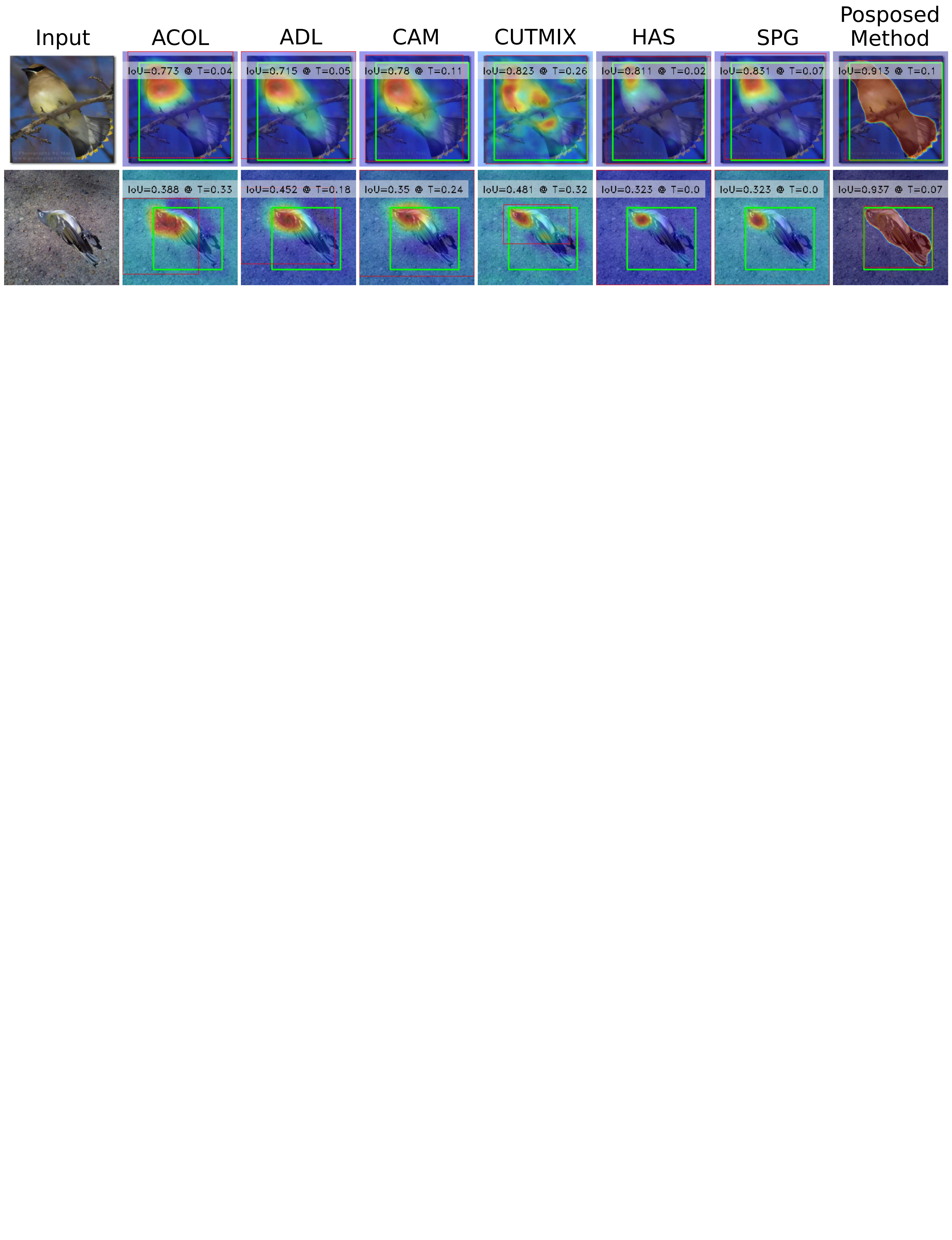}
		\vspace{-1.7em}
		\captionof{figure}{Samples from CUB test set.}
		\label{fig:results_cub}
\end{figure*}

To validate our proposed model, we employed the well-known CUB-200-2011 dataset for object localization. Following the protocol in \citep{choe2020evaluating}, we trained our model using an independent validation set, and a new standard metric named \newmaxboxacc. Results shown in Table \ref{tab:cub_maxbox_v2} and Fig. \ref{fig:results_cub} indicate that our proposed approach can outperforms related state-of-art methods. We also compare the activation distribution of our method with baseline methods in Fig \ref{fig:dist_shift} that shows baseline methods have very close pixel values for foreground and background regions, while our proposed method can distinguish foreground and background more efficiently, eliminating the need of threshold per image.

\vspace{-1.1em}
\begin{table*}[!htp]
\centering
\resizebox{\linewidth}{!}{%
\centering
\small
\begin{tabular}{lc*{3}{c}gc*{3}{c}gc*{3}{c}gc*{3}{c}gc*{3}{c}gc*{3}{c}g}

& &  \multicolumn{4}{c}{\textbf{\maxboxacc}} & & \multicolumn{4}{c}{\textbf{\newmaxboxacc}}  \\
\cline{3-6}\cline{8-11} \cline{3-6}\cline{8-11}  
Methods   &  & VGG & Inception & ResNet & Mean &  & VGG & Inception & ResNet & Mean  \\
\cline{1-1}\cline{3-6}\cline{8-11}
CAM~\citep{zhou2016learning} {\small \emph{(cvpr,2016)}} &  & 71.1 & 62.1 & 73.2 & 68.8  &  & 63.7 & 56.7 & 63.0 & 61.1 \\
HaS~\citep{singh2017hide} {\small \emph{(iccv,2017)}} &  &         76.3 & 57.7 & 78.1 & 70.7 &  &        63.7 & 53.4 & 64.7 & 60.6 \\
ACoL~\citep{zhang2018adversarial} {\small \emph{(cvpr,2018)}} &  &      72.3 & 59.6 & 72.7 & 68.2 &  &      57.4 & 56.2 & 66.5 & 60.0 \\
SPG~\citep{zhang2018self} {\small \emph{(eccv,2018)}} &  &       63.7 & 62.8 & 71.4 & 66.0 &  &       56.3 & 55.9 & 60.4 & 57.5 \\
ADL~\citep{choe2019attention} {\small \emph{(cvpr,2019)}} &  &           75.7 & 63.4 & 73.5 & 70.8  &  &           66.3 & 58.8 & 58.4 & 61.1  \\
CutMix~\citep{yun2019cutmix} {\small \emph{(eccv,2019)}} &  &     71.9 & 65.5 & 67.8 & 68.4 &  &     62.3 & 57.5 & 62.8 & 60.8 \\
F-CAM \citep{belharbi2022f} {\small \emph{(wacv,2022)}} &  &    91.5 & 86.8 & 92.4 & 89.7 &  & 84.3 & 76.2 & 82.7 & 80.3 \\
\cline{1-1}\cline{3-6}\cline{8-11}
\textbf{Proposed Method} &  &                               \multicolumn{3}{c}{\textbf{97.0}} & -- &  & \multicolumn{3}{c}{\textbf{90.9}} & -- \\
\cline{1-1}\cline{3-6}\cline{8-11} \\
\end{tabular}
}
% \vspace{0.6em}
% \vspace{-1em}
\caption{Localization performance on CUB-200-2011.}
\label{tab:cub_maxbox_v2}
\vspace{-1em}
\end{table*}

\section*{Acknowledgment}
This research was supported in part by MITACS, the Ericsson Global AI Accelerator Montreal, and Compute Canada.

% \clearpage
% \newpage

\appendices

% ===================================================================================================
% 
%
%                                          SUPP-MATERIAL
% 
% 
% ===================================================================================================

\bibliographystyle{apalike}
\bibliography{biblio}

\end{document}

%% file: macros.tex
\usepackage{times}
\usepackage[numbers]{natbib}
\usepackage[english]{babel}
\usepackage{blindtext}
\usepackage{graphicx}
\usepackage{amsmath, amsthm, amssymb, bbm, bm}
\usepackage{enumerate}
\usepackage{float}      % figure inside minipage,
\usepackage{subcaption}  % does not go well with subfig.
\usepackage{wrapfig}
\usepackage[margin=0cm]{caption}
\usepackage[titletoc, toc]{appendix}
\usepackage{tabularx}

\usepackage{multirow}
\usepackage{hhline}
\usepackage{makecell}
\usepackage{placeins}  % stop floating

\usepackage[x11names, usenames, dvipsnames, svgnames, table]{xcolor}
\definecolor{firebrick}{rgb}{.698,.133,.133}
\definecolor{mybluelight}{rgb}{0.9, 0.9, 1.}
\definecolor{myorangelight}{rgb}{1., 0.9, 0.9}

\usepackage[utf8]{inputenc} % allow utf-8 input
\usepackage[T1]{fontenc}    % use 8-bit T1 fonts
\usepackage{url}            % simple URL typesetting
\usepackage{booktabs, colortbl}       % professional-quality tables
\usepackage{amsfonts}       % blackboard math symbols
\usepackage{nicefrac}       % compact symbols for 1/2, etc.
\usepackage{microtype}      % microtypography

\usepackage{csquotes}
\usepackage{latexsym}

\usepackage{pifont}% http://ctan.org/pkg/pifont
\usepackage[boxruled, vlined, linesnumbered]{algorithm2e}
\SetAlFnt{\small}
\SetAlCapFnt{\small}
\SetAlCapNameFnt{\small}
\usepackage{algorithmic}
\algsetup{linenosize=\tiny}

\let\oldnl\nl% Store \nl in \oldnl
\newcommand{\nonl}{\renewcommand{\nl}{\let\nl\oldnl}}% Remove line number for one line

\usepackage{paralist}

\usepackage{xspace}
\usepackage{soul}
\usepackage{dsfont}
\usepackage{stmaryrd}
\usepackage[textwidth=15mm]{todonotes}
\usepackage{dirtytalk}
\usepackage{pbox}
\usepackage{cprotect}

\usepackage{verbatim}
\usepackage{textcomp}
\usepackage[normalem]{ulem}

\usepackage{mathtools}
\usepackage{etextools}
\usepackage[inline]{enumitem}

\usepackage[colorlinks=true,allcolors=firebrick,bookmarks=false]{hyperref}

\newcommand\maxboxacc{\texttt{MaxBoxAcc}\xspace}
\newcommand\newmaxboxacc{\texttt{MaxBoxAccV2}\xspace}

\definecolor{darkergreen}{RGB}{21, 152, 56}
\definecolor{red2}{RGB}{252, 54, 65}
\definecolor{Gray}{gray}{0.85}
\newcolumntype{g}{>{\columncolor{Gray}}c}

\let\OLDthebibliography\thebibliography
\renewcommand\thebibliography[1]{
  \OLDthebibliography{#1}
  \setlength{\parskip}{0pt}
  \setlength{\itemsep}{0pt plus 0.3ex}
}
\theoremstyle{definition}

\DeclarePairedDelimiterX{\divx}[2]{(}{)}{%
  #1\;\delimsize\|\;#2%
}